\RequirePackage{amsthm}
\documentclass[sn-basic, pdflatex]{sn-jnl}
\usepackage{graphicx}\usepackage{multirow}
\usepackage{amsmath,amssymb,amsfonts}
\usepackage{mathrsfs}
\usepackage[title]{appendix}\usepackage{xcolor}\usepackage{textcomp}\usepackage{manyfoot}\usepackage{booktabs}\usepackage{algorithm}\usepackage{algorithmicx}\usepackage{algpseudocode}\usepackage{listings}\usepackage{enumerate}
\usepackage{tablefootnote}
\usepackage{comment}
\usepackage{soul}

\usepackage{color, soul}
\renewcommand{\hl}{}
\theoremstyle{thmstyleone}
\theoremstyle{thmstyletwo}
\theoremstyle{thmstylethree}
\begin{document}
\title[Article Title]{
\hl{
Multimodal Reaching-Position Prediction for ADL Support Using Neural Networks
}
}

\author*[1]{\fnm{Yutaka} \sur{Takase}}\email{yutaka\_takase@shinshu-u.ac.jp}
\author[1]{\fnm{Kimitoshi} \sur{Yamazaki}}\email{kyamazaki@shinshu-u.ac.jp}

\affil[1]{\orgdiv{Mechanical System Engnieering}, \orgname{Shinshu University}, \orgaddress{\street{Wakasato 4-17-1}, \city{Nagano}, \postcode{3808553}, \state{Nagano}, \country{Japan}}}

\abstract{
  This study aimed to develop daily living support robots for patients with hemiplegia and the elderly. To support the daily living activities using robots in ordinary households without imposing physical and mental burdens on users, the system must detect the actions of the user and move appropriately according to their motions.
  We propose a reaching-position prediction scheme that targets the motion of lifting the upper arm, which is burdensome for patients with hemiplegia and the elderly in daily living activities.
  For this motion, it is difficult to obtain effective features to create a prediction model in environments where large-scale sensor system installation is not feasible and the motion time is short.
  We performed motion-collection experiments, revealed the features of the target motion and built a prediction model using the multimodal motion features and deep learning.
  The proposed model achieved an accuracy of 93 \% macro average and F1-score of 0.69 for a 9-class classification prediction at 35\% of the motion completion.
}

\keywords{reaching-position prediction, multimodal networks, daily living support}
\maketitle

\section{Introduction} \label{sec:introduction}
With an aging society, the demand for intelligent robots to support activities of daily life (ADL) for elderly and disabled people living alone is increasing.
These robots are required to work close to human users in environments that are relatively narrow and difficult to sense, in contrast with industrial robots, which work in well-controlled environments of factories or warehouses.
The influence of the support provided by the autonomous robot on the mental health of the user must also be considered.
For example, is not always appropriate for the assisting robot to fully support a user who intends to reach a distant object by picking up the object and bringing it to the user. 

In this case, despite the intention of users to move on their own, the support ends up undermining the intention. Therefore, it is essential for the ADL  supporting system not to inhibit the active motivation and self-efficacy of users (\cite{Bandura1978-dq, Stewart2019-mi}. In \cite{Stewart2019-mi}), the authors reported a correlation between SE evaluation scores and movement speed and accuracy for reaching motions of patients with residuals from stroke.

This study aims to develop an autonomous ADL support robot, considering the intentions of the user. We focus on cooperation task completion with users to maintain and improve their self-efficacy. In the previous example, the robot could support the arm of the user to reach out to pick up the object or move the target object into an easier-to-pick position. Such a support system, not only maintains the self-efficacy of the users but also improves it through the experience of accomplishing tasks that would be difficult to accomplish alone.

To achieve the goals of the study, in this paper, \hl{we propose a novel scheme to predict reaching position in reaching motion involving upper-arm lifting.}
Although lifting the upper arm is an essential part of ADL, such as picking up an object from a high place, putting it up, or drying laundry, it is difficult for elderly or disabled patients because of the need to maintain their arms at high positions, which may cause an imbalance in the torso.
There are many possible actions that the assisting robot can perform to support the motions, such as directly supporting the arms and torso of the user and interfering with objects that are the target of movements of the user.

Therefore, establishing a prediction scheme for this motion along with an analysis of the features of the motion will be useful for both hardware and software development of support robots.

The main contributions are summarized as follows:

\begin{itemize}
\item We collected the motion data of lifting upper arm, which imitates object grabbing using multiple sensors, and analyzed the motion features.
\item We created a multimodal-neural-network model to predict the reaching position as a classification problems that can be adapted to real-time robot control.
\end{itemize}
The remainder of this paper is structured as follows: In the next section, we introduce related works. In Section \ref{sec:approach}, we defined our research problems and approaches. In Section \ref{sec:data_collection}, we describe the motion data collection and analysis. Then, we proposed our multimodal reaching-position prediction network in Section \ref{sec:prediction_model}. The results and discussion are presented in Sections \ref{sec:result} and \ref{sec:discussion}.  Finally, Section \ref{sec:conclusion} concludes the paper.

\section{Related Work} \label{sec:related_works}
There are various approaches to developing cooperative robots.
Research themes in this area are shifting from proposing robot motion generation methods to developing human-motion estimation frameworks.
For tasks where robots and humans work in proximity, as in this study, there are systems aimed at sharing the workspace to avoid interference with each other, and systems aimed at cooperating when performing a single task, like in handover or load-sharing tasks.

Assembly tasks at a factory are typical scenes of cooperative tasks between robots and humans (e.g. \cite{Arai2020-ec, Mainprice2015-vt}).
In the case of a robot and humans sharing a workspace and working individually, the robot must predict their motion trajectory to avoid collisions with humans. In \cite{Arai2020-ec}, the authors addressed the problem of estimating the reaching motion of human workers as a multi-class classification problem. They reported that the accuracy of the proposed method, which used 3D point-cloud data, was around 80 \% after 50\% of the operation. In \cite{Mainprice2015-vt}, the authors collected data on the reaching motions of workers using a motion capture system and used the data to predict the arm trajectory of human worker. The construction of advanced sensor environments is beneficial in factory and laboratory environments.

In research on handover tasks, which require the positions of robots and humans to be close, sensors such as voice and electromyography sensors are used to predict the trajectory of a workers' arms (\cite{Wang2021-qe, Wu2019-wb}).

In the area of human-robot interaction, many studies have been constructed models to predict user activities and intentions using various features as well as the movement of the users to achieve a natural interaction between humans and robots that is similar to human and human interaction. 
For example, in \cite{Tan2019-qc,Zadeh2016-rk}, the authors proposed emotion estimation models using the facial expressions and verbal features of the user. In \cite{Wang2021-qe}, the authors proposed a deep-neural-network model to estimate the order of the users for the robot by using both verbal and nonverbal features. In \cite{Yuguchi2019-iw}, the user intention to service robots was estimated using facial direction. 
Although using human natural motions, such as facial direction, seems effective in informing the intention or purpose of the motion to systems, the system we aim for, as described above, targets supporting daily life activities according to the actions of the user. Therefore, it is not appropriate to build complex sensor systems for trajectory tracking in the home or give voice instructions to robots like "I want to get the book on the upper right shelf."
In this study, we address these problems by using simple sensor systems.
In addition, we deal with the reaching-position prediction problem for upper-arm lifting motion as a multi-class classification task and create a novel model that uses multi-nonverbal features.
The proposed method could be applied in the future to load-sharing tasks (\cite{DelPreto2019-cf, Sirintuna2022-vv}. Currently, studies focus on control methods and algorithms after humans and robots hold the load; however, this study proposes one approach to the important problem of how a robot can hold objects together according to human intentions.

In the following section, we present our research questions and approaches.

\section{Research questions and Our approaches} \label{sec:approach}
\subsection{Research questions}
This study addressed two research questions: First, we investigated the practical features of upper-arm lifting motion to construct a reaching-position prediction model; second, we built a neural-network model using the features.
Considering our goal, we assumed the following use environment and scenario: The system would be used in the everyday household environment, the users would be patients with hemiplegic and older adults with weakened muscles, and the support system should operate autonomously, and avoiding compromising the self-efficacy of the user by not providing full support.

For a specific task, assume that the user takes an object from a shelf with a healthy arm. The system recognizes the reaching position of the motion and interacts with the user's arm and the object to be grabbed. This means that the system supports the task by, for example, keeping the arm or torso or moving the object to a position that is easier to grasp.

Based on these assumptions, the proposed method has the following requirements.
\begin{itemize}
  \item It deals only with available information without installing or attaching large sensor systems to the user or environment.
  \item The proposed method assumes that the support system works autonomously without active manipulation by the user for operation. 
  \item It provides an environment in which the user does not have to wait for support or adjust their operating speed.
\end{itemize}

\subsection{Approach}
Under the conditions described above, our approach to investigating the research question is as follows.
First, we collect target motion data of multiple subjects in an assumed environment.
Next, the features of the motions are selected from the collected data, which are considered adequate for constructing a prediction model.
Finally, as in \cite{Arai2020-ec}, we constructed a prediction model of the reaching position as a multi-class classification problem using deep learning and evaluated its performance.

The next section describes the data collection method, its features, and the features that can be used to predict the arrival position.
\section{Analysis of reaching motions} \label{sec:data_collection}
\begin{figure}[ht]
   \centering
   \includegraphics[scale=0.5]{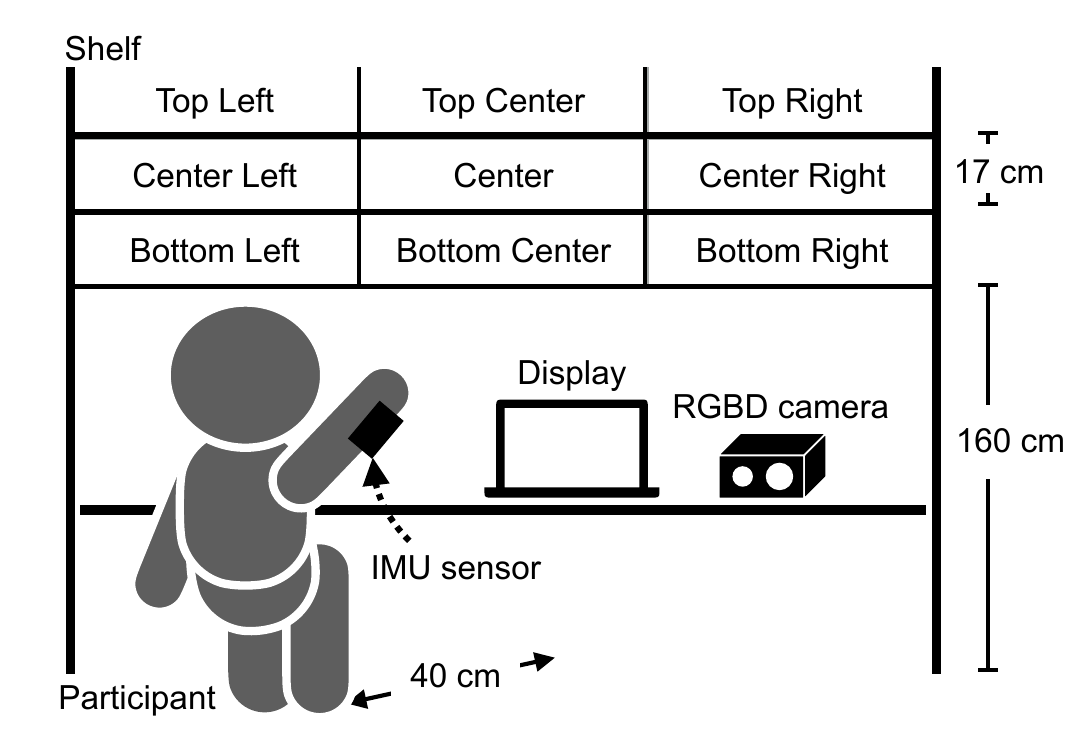} 
   \caption{Overview of data collection environment}
   \label{fig:shelf}
\end{figure}

\subsection{Motion collection} \label{sec:motion_collection}
\figurename{ \ref{fig:shelf}} shows the environment settings for the data collection. The motion data collection procedure is as follows.
As illustrated in the figure, the participant sat on a chair in front of a shelf divided into nine regions. The participant performed the motion of grabbing things from the area randomly indicated by the experimenter. Every indication was presented visually after a 3-second countdown in one region to be determined instantly on a display set in front of the participant. One set of trials consisted of four randomly ordered motions to each region repeated four times, and all participants sequentially performed seven sets of trials.
The participants were instructed to place their right hands on their knees and face the display in front of them during the countdown.
The sensors used were an RGBD camera (Microsoft, Azure Kinect) installed in front of the participant and an inertial measurement unit (IMU) sensor (MicroStrain, 3DM-Gx5-45) attached to the right arm of participants. \hl{Color (resolution: 1280 $\times$ 720, field of view: 90$^{\circ}$ $\times$ 59$^{\circ}$) and depth images (resolution: 640 $\times$ 576, field of view: 75$^{\circ}$ $\times$ 65$^{\circ}$) were acquired from the RGBD sensor at 15 frames per seconds (FPS), and magnetometer, angular velocity, and acceleration data were obtained from the IMU sensor at 100 FPS.}

\begin{figure*}[t]
    \centering
    \includegraphics[scale=0.25]{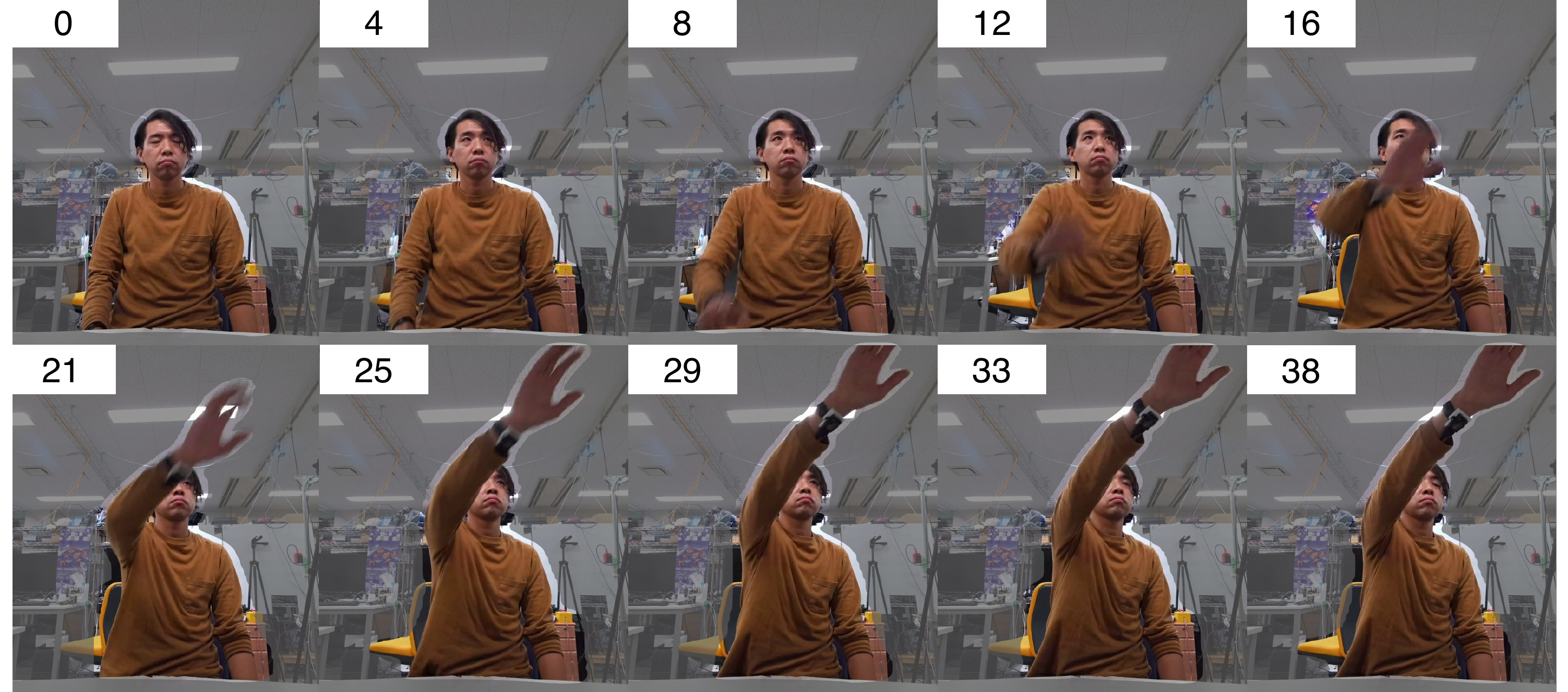}
    \caption{Example of the collected motion data; it is a sequence of color images subject to reach the center-left region. The number indicates elapsed frames from the start of the motion.}
    \label{fig:motion_exp}
\end{figure*}

\hl{Six able-bodied male participants (aged 22-25, all right-handed)} were recruited from our laboratory. Excluding data recording failures, the effective number of data samples was 1538.
\figurename{ \ref{fig:motion_exp}} shows an example of the collected data: the sequence of reaching motion to the center-left region. The numbers indicate the elapsed frames from the start of the motion.
\subsection{Motion analysis}\label{sec:motion_analysis}

\begin{table}[t]
\centering
\caption{\hl{Descriptive statistics of the reaching motions time(s) by target region}}
\begin{tabular}{rrrrrrr}
\toprule
Region & N & Mean & Median &  Max & Min & SD\\
\midrule
Top-left    (TL) & 173 & 1.56 & 1.56 & 2.20 & 1.10 & 0.230 \\
Top-center   (TC)& 169 & 1.56 & 1.50 & 2.28 & 0.969& 0.295 \\
Top-Right    (TR)& 171 & 1.58 & 1.56 & 2.43 & 1.07 & 0.267 \\
Center-left  (CL)& 174 & 1.45 & 1.46 & 2.77 & 1.00 & 0.260 \\ 
Center       (C)& 172 & 1.42 & 1.37 & 2.10 & 0.900& 0.242 \\
Center-right (CR)& 170 & 1.48 & 1.47 & 2.38 & 0.902& 0.289 \\
Bottom-left   (BL)& 171 & 1.44 & 1.45 & 1.97 & 0.883& 0.219 \\
Bottom-center (BC)& 170 & 1.33 & 1.33 & 1.93 & 0.868& 0.227 \\
Bottom-right (BR)& 170 & 1.37 & 1.35 & 2.57 & 0.732& 0.285 \\
\botrule
\end{tabular}
\label{tab:reaching_time}
\end{table}

\hl{\mbox{Table \ref{tab:reaching_time}.} shows the descriptive statistics for the reaching times to each region derived from the collected data. Here, the reaching time is measured based on video data, from the moment the participant starts the movement after the target region is indicated by the display to when the extended right arm becomes stationary. Therefore, the time it takes for the visual reaction is not included.
One-way ANOVA and the post hoc Tukey HSD test ( $p <.05$ was considered significant ) were conducted and suggested that there were significant differences between the regions (F(8, 1529) = 19.87, $p < .001$). All the post hoc test results are shown in \mbox{\figurename{ \ref{fig:result}}}, which will be discussed later.
From the results, reaching the uppermost regions, which are the farthest from the right hand’s initial position, top-left (TL), top-center (TC), and top-right (TR), required approximately 1.56, 1.56, and 1.58 s, respectively. And, there was no significant difference observed between them. 
Similarly, no significant difference was observed among the middle regions, center-left (CL), center (C), and center-right (CR). These results suggest that, within this experimental setup, participants unconsciously adjust their movement speed to reach regions at the same height. This adjustment equals modulating the waiting time until the next movement's target position is presented. Therefore, it might be influenced by experimental conditions.
On the other hand, for the bottom regions, reaching the bottom-center (BC), which is located /directly in front of the body, was the fastest, with an average time of about 1.33 s. According to the post hoc test results, there was no significant difference between BC and bottom-right (BR). However, significant differences ($p <.05$) were observed between BC and the bottom-left (BL), where required to extend the right arm to the front-left. This suggests that such a movement appears to be particularly difficult, even for healthy individuals.
Additionally, it was observed that the maximum reaching time to BR was relatively larger compared to the other bottom regions. Upon reviewing the video data of this motion, it was noted that, after quickly getting closer to the target location, participants continued a slow approach movement without coming to a complete stop. This data has not been excluded, as it is considered not to affect future analyses or system development significantly.
Thus, in simple reaching movements, while there is an observed tendency to unconsciously adjust speed, it became clear that due to presence of locations significantly more difficult to reach, reaching speed and movements are not solely determined by simple distance between the arm and the destination.}

This study uses the average value of 1.47 s from all data as a guideline for developing a system to support this task. The proposed system must perform user motion recognition, predict the reaching position, and provide support actions all within this time frame.

\begin{figure}
    \centering
    \includegraphics[width=0.75\linewidth]{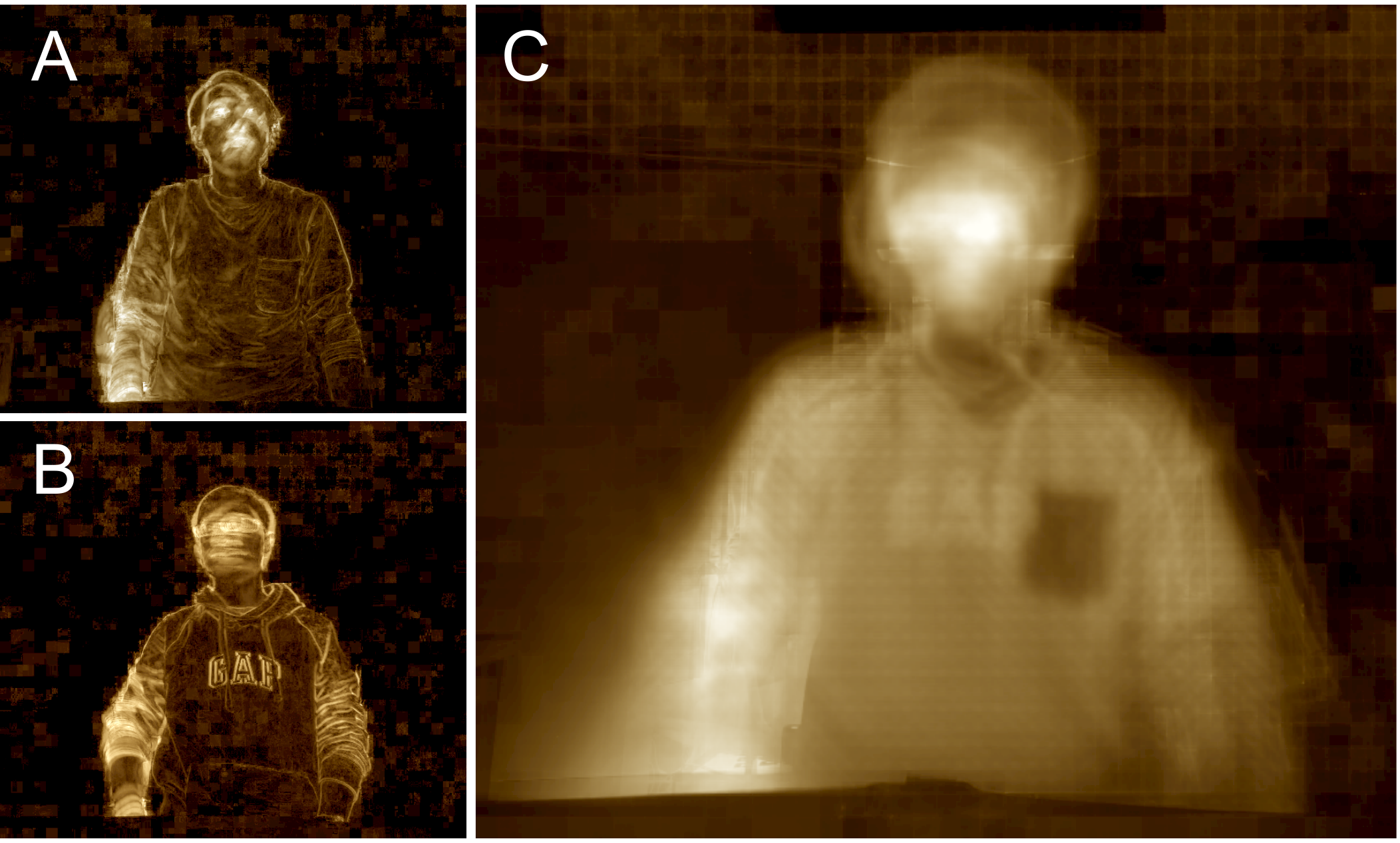}
    \caption{\hl{Visualization of variations within 10 frames after the start of the motion using the sum of absolute difference (SAD). A, B: Examples from the collected motions. C : Result based on the entire collected motions. }}
    \label{fig:motion_diff}
\end{figure}

\hl{Additionally, \mbox{\figurename{ \ref{fig:motion_diff}}} shows the differential images created using the SAD (Sum of Absolute Differences) method from color images over 10 frames after the start of the motion. {\ref{fig:motion_diff}}A and {\ref{fig:motion_diff}}B are from specific motions extracted from the collected dataset, while {\ref{fig:motion_diff}}C is generated from all data. Bright areas in the images indicate regions of significant movement within the frames.
The collected video data and the differential image also revealed the following features.

1) The target motion consisted of movements of specific body parts, that is, upper body, right arm, and face, rather than the entire body; in particular, from the differential image, the upper body movements appear not to be significant in the initial phase of the motion, making it seem challenging to use this information to predict the reaching position.
On the other hand, the image shows significant movement near the face and around the right arm.
Therefore, it was considered effective to use the right arm motion with visual features and to capture changes from the early stage of the motion together with face direction changes.
}

2) The preparatory motion was not useful for prediction; in other words, the time used to predict from the start of the motions directly affected the time available for the support action, and it was also necessary to recognize the timing of the start of the motion. However, it was challenging to obtain the exact timing of the start of the motion due to there was no prior motions. This point is discussed in Section \ref{sec:discussion} as a future issue.
\subsection{Modals for prediction}
Based on the observation results stated above, we selected the face, visual, and motion features to construct a reaching-position prediction model.
The observation results and \cite{Yuguchi2019-iw} suggested that the face direction or features would be an essential cue to estimate the following motions.
Additionally, to use this system universally, it is more appropriate to estimate the reaching position using depth information as visual cues rather than color image data, which contains redundant external information related to the user and the environment.
Furthermore, motion features acquired using the IMU sensor attached to the healthy side wrist of the user were employed,
\hl{because, considering future robots supporting the user’s arms or grasping objects, it becomes imperative to understand the three-dimensional movements and postures of the arm. 
}
Although there are many technologies to estimate human postures by using only color images \cite{Cao2021-oa}, considering the typical house environment in Japan,  it is difficult to obtain a camera angle of view sufficient to estimate the arm's posture in reaching an unspecified direction to a shelf placed in front of the user. Therefore, estimated posture data were not used.
Eye-tracking devices were also not used to avoid complicating the system.
\subsection{Motion data extraction}
In this study, the start and end recognition of the motion was not performed. Therefore, it was necessary to extract data of each motion from the collected data based on some criteria. We manually annotated the motion start and end timings based on the following definition: The motion-start timing was defined as the frame when the right hand, initially positioned at the knee, began to move. The motion-end timing was determined as a frame when the extended arm started to retract at the reaching position. The collected data were divided into individual motions according to the annotated timings.

In the next section, we organize the features discussed in this section into specific feature data and discuss the construction of the reaching-position prediction model.

\section{Prediction Model} \label{sec:prediction_model}
\subsection{Features}
This section describes the features used to build our prediction model.
\subsubsection{Face Features}
\hl{
Face feature was set with the expectation of capturing the direction of the face, its variations, and the characteristics of gaze transition.
In studies to create prediction models of human behavior, movements of the head and gaze are often used as features \mbox{(\cite{Holman2021-eo, Yang2023-gx}).}  We attempted to capture such features without attaching sensors to the users.} In this study, the face mesh data was employed as face features. We used Google Mediapipe (\cite{Lugaresi2019-wm}) to obtain  468 3D face landmark positions from the color images. The time elapsed from the start time of the motion was added to each frame, resulting in data of 1405 (468 $\times$ 3 + 1) dimensions.
\subsubsection{Depth Features}
\hl{
Depth features were set with the expectation of extracting the three-dimensional characteristics of movements while reducing dependency on clothing and the experimental environment. These features are valuable for understanding the user's position and posture in future support scenarios.
}
The depth image data was acquired at 15 FPS and cropped to the user center. The resolution was reduced to 256 $\times$ 188. The elapsed time was also added to each frame, as described below.
\subsubsection{Motion Features}
\hl{Motion features were set to capture the characteristics of rapid three-dimensional movements of the arm.}
As the motion features, the data from the IMU sensor attached to the right wrist provided ten dimensions of information (geomagnetism, acceleration, and angular acceleration). The elapsed time was also added to the data to obtain 11 dimensions.

\subsection{Network structure}
\begin{figure}
    \centering
    \includegraphics[scale=0.32]{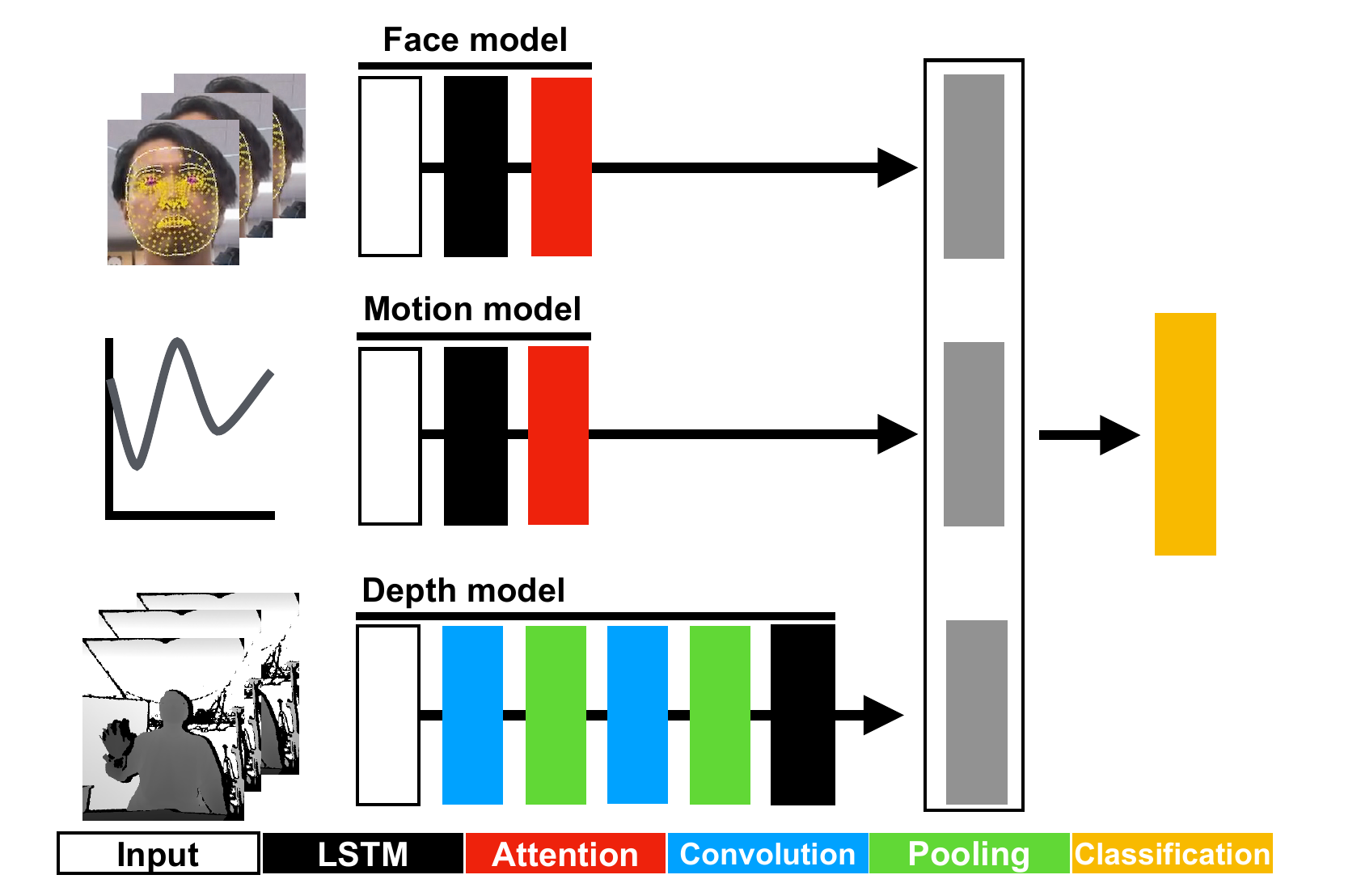}
    \caption{Multimodal late fusion model}
    \label{fig:model}
\end{figure}

We constructed a multimodal 9-class classification neural-network model to predict reaching positions as seen in Section \ref{sec:data_collection}.
The long short-term memory (LSTM) (\cite{Hochreiter1997-hk}), and the local attention mechanism (\cite{Vaswani2017-go}) were used to construct our machine-learning model. The network structure is shown in \figurename{ \ref{fig:model}}.
As shown in the figure, the output from all modal layers are combined through late fusion (\cite{Gadzicki2020-jh,Sun2021-rl}).
This is the method used in modeling multimodal information. The composition of each unimodal network is as follows.
\subsubsection{Face layers}
A bi-directional LSTM layer with 1405-dimensional input and 2048-dimensional output was employed to train the face modal. The final output data passed through a self-attention layer and was output as 2048-dimensional data. The number of parameters in the network was 44,554,241.

\subsubsection{Motion layers}
The same structure as the face model was used to train the motion model. 
The number of input dimensions was 11, and the number of output dimensions was 512. It had 2,139,137 network parameters.
\subsubsection{Depth layers}
Depth features were learned by combining latent representation of depth images using a convolutional neural network (CNN) and time-series learning using LSTM (\cite{Mou2021-qo}). The CNN parameters used are shown in Table \ref{tab:cnn_parameters}.
The CNN + LSTM network had a total of 204,509,720 parameters, and the latent representation of each frame of the CNN data was combined with the elapsed time, described earlier, and input to the LSTM layer.
\begin{table}[t]
\centering
\caption{CNN model parameters for depth network}
\begin{tabular}{r r }
\hline
1D CNN & \\
\hline
Convolutional layer & Filter=8, kernel-size=(16,16),stride=1\\
Max-pooling + Dropout & pool-size =(2,3)\\
Convolutional layer& Filter=16, kernel-size=(8,8),stride=1\\
\\
Max-pooling + Dropout & pool-size =(8,8)\\
Fully connected layer & input-size= 25088, output-size= 2047\\
\hline
\end{tabular}
\label{tab:cnn_parameters}
\end{table}

\subsubsection{Classification layers}
The three output vectors from the unimodal layer were simply combined into a 1 $\times$ 4097-dimensional vector. They were then input to a fully connected layer with dropout at each layer.
The dimension of the last output layer was nine, the number of classes.
The dropout rates were set to 0.6, 0.4, and 0.2, respectively, and the ReLU function was used as the activation function.

\subsection{Input frames}
As shown in Table \ref{tab:reaching_time}, the target motion time for this task was a minimum of aproximately 1.33 s to complete the motion. Even if we disregard the movement time of the support robot, we still need a prediction of a shorter time to assist the robot's movement.
The time used for the prediction was set to 0.5 s (7 frames for the face and depth models and 50 frames for motion model). This means the model used information from 32\% to 36\% of the motion time. This is a short prediction time compared to the previous study (\cite{Arai2020-ec}).
To improve the performance of the prediction model, data interpolation should be performed for missing data. However, considering real-time use, the raw data obtained from the sensors should be input to the predictor with as little processing as possible. Therefore, data shaping was kept to a minimum. For example, frames where face mesh could not be recognized were padded with zeros.

In the next section, we discuss  the result of the training and the features of our prediction model.

\section{Evaluation and Result}\label{sec:result}
\subsection{Model performance}
\begin{table}[t]
\centering
\caption{Single vs Multi modal model performance comparison}
\begin{tabular}{r r r r r}
\hline\hline
Model & Accuracy & F1-score & Precision & Recall\\
\hline
face Model &0.91& 0.59 &0.60& 0.59 \\
imu Model &\textbf{0.93} & 0.66 & 0.66 & 0.68\\
depth Model &0.92 & 0.64 & 0.65& 0.64\\
Fusion Model&\textbf{0.93}&\textbf{0.69}&\textbf{0.69}&\textbf{0.69}\\
\hline
\end{tabular}
\label{tab:comp_result}
\end{table}

\begin{figure}[t]
\centering
\includegraphics[scale=0.26]{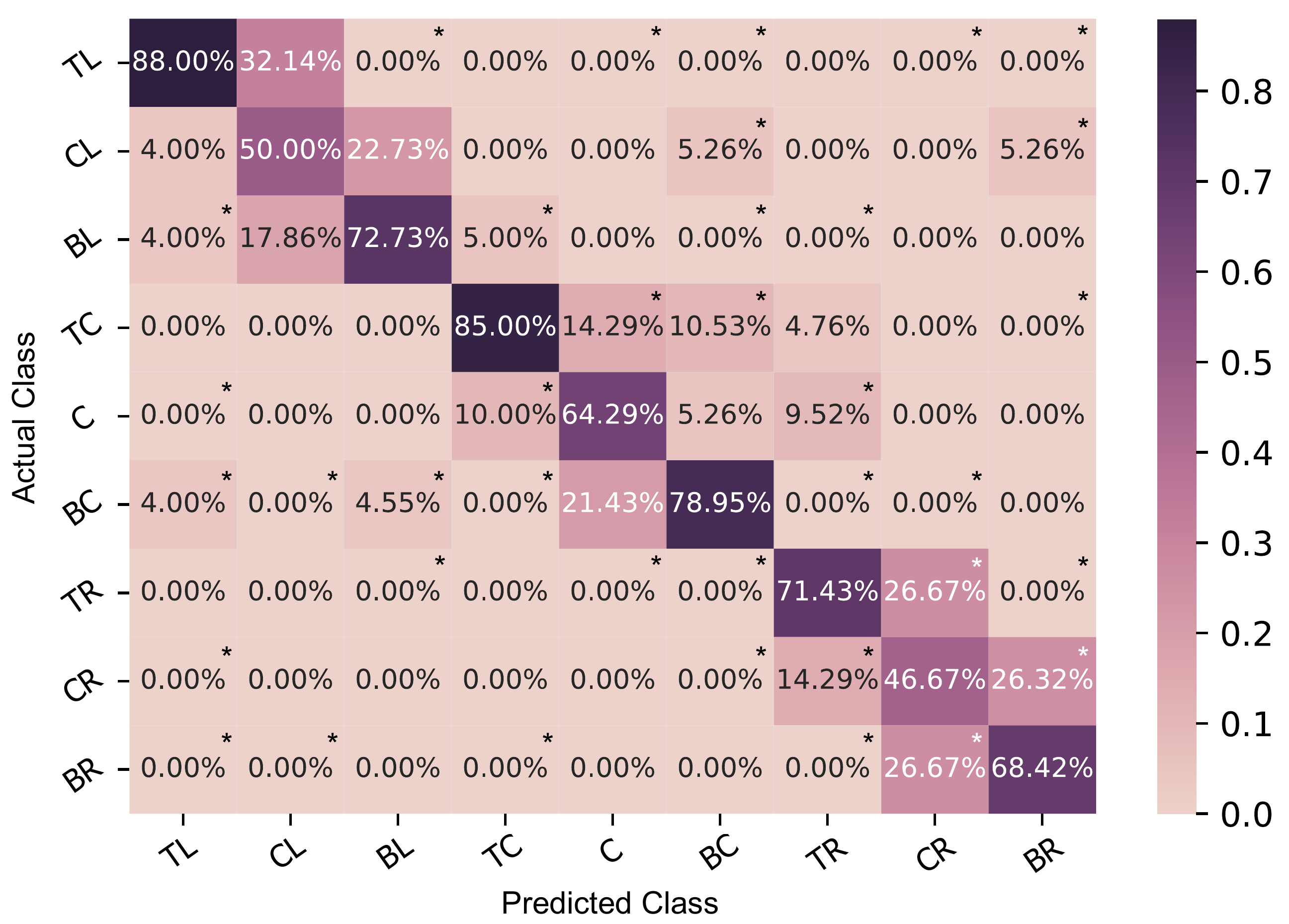}
\caption{\hl{Confusion matrix of the proposed fusion model. the percentages represent classification rate.The asterisks indicate pairs for which a significant difference in motion speed was found as a result of the post-hoc Tukey HSD test describe in Section \mbox{\ref{sec:motion_analysis}}.}}
\label{fig:result}
\end{figure}
For training and evaluation the proposed late fusion model, firstly, the whole dataset is randomly divided into the training set and test set with 1341 and 197 motions, respectively. Then, we trained the model using a 10-fold cross-validation strategy. \hl{Table \mbox{\ref{tab:comp_result}}.
shows model accuracy and Macro Precision, Recall, and F1-score, which are used to evaluate multi-class classification problems, of the fusion model and each unimodal model for comparison. These values are obtained by calculating the scores for each class in a one-vs-rest manner and then averaging these results across all classes.}
The unimodal models were trained only using LSTM (CNN-LSTM) and a classifier structure for each model in the \figurename{ \ref{fig:model}}.
The results found that the proposed model performs well or better than other unimodal models. In particular, the fusion model has the highest F1-score of 0.69, indicating that it was the most balanced model.
In a previous study addressing a similar 9-class reaching-position prediction problem, an accuracy of 80\% was reported at approximately 50\% of the motion completion time. In contrast, our method achieved higher accuracy (93 \%) at an earlier stage of the motion (32\%).
\hl{ \mbox{\figurename{ \ref{fig:result}}} shows a confusion matrix obtained from 197 test data. The rows represent the actual classes and columns represent the predicted classes by the fusion model. The percentages represent the precision of the classification results for each class. For example, data classified as TL are correctly classified with an 88.0\% probability; however, it shows a 4.0\% probability of ML, BL, and BC motions being incorrectly classified as TL.
The results show that the precision increases from the right side, near the starting point of motion, towards the upper left. Also, as a trend across the classes, there is some confusion within the same column. 
In particular, the precision for the middle row was relatively low, often confused with motions to the same column. This is an interesting result, suggesting that even from the initial few frames, movements towards the top and bottom regions are distinctively characteristic. Distinguishing whether the arm stops in the middle row or continues moving up or down is difficult for the current model. 
However, the results from the post-hoc test, conducted in Section \mbox{\ref{sec:data_collection}}, also show potential for classification. The asterisks in the figure indicate pairs where a $ p<.05$ significant difference in motion speed was observed in the post-hoc Tukey HSD test.
From the result, significant differences were observed in the final reaching motion time between CR and TR, and CR and BR, indicating that there are differences in arm motions between them.
Even if there are differences in the current input frames regarding movement trajectories, this model, which predicts reaching positions by combining arm, face, and depth features, has cases where classification may fail due to factors other than motion features.
To analyze the factors causing differences in reaching times, arm trajectories and features must be analyzed, and neural networks capable of extracting these data must be constructed. These tasks remain for future research.
Additionally, it is impossible to avoid the possibility of misclassification completely.
Based on these results, the operation of future support systems will be discussed in Section \mbox{\ref{sec:discussion}}.
}
\subsection{Estimation speed} \label{sec:estimation_speed}

Finally, we measured the estimation speed of the proposed model --- with 282,930,211 parameters. The computer used for inference was Ubuntu 20.04.6 LTS for OS, Intel(R) Xeon(R) W-2225 @ 4.10GH CPU, NVIDIA RTX A5000 and 24GB RAM.
\hl{
The input data utilized the motion data collected in Section \mbox{\ref{sec:data_collection}}. These data are stored using the functionalities of Robot Operating System \mbox{\cite{Quigley2009-dc}}, allowing for the simulation of receiving camera images and IMU sensor data while maintaining timestamp information. However, delays such as the camera's image acquisition time or data transmission between the sensor and the computer are not considered.
The measurement program measured the time from when it collected the number of input frames of sensor data through the conversion and trimming into a format suitable for the predictive model to obtain the prediction results. The prediction model is trained using PyTorch \footnote{PyTorch : \url{https://pytorch.org}} and optimized with Nvidia TensorRT \footnote{Nvidia TensorRT : \url{https://developer.nvidia.com/tensorrt}}. 
}
\hl{The average prediction time for 100 data inputs was 0.0086 s, with a standard deviation of 0.0036 s. The maximum value was 0.022 s, and even if this worst-case scenario is adopted, the time required for estimation is approximately 1.5 \% of the motion time of our collected motion data, which is considered sufficiently small.}
This result indicates that the proposed method requires a prediction time of approximately 0.5 s (for collecting motion data) + 0.086 s (for estimating the target position) for a reaching motion that takes approximately 1.47 s on average.
Therefore, the proposed system leaves appropriately 0.96 s of grace time for the support robot.

\section{Limitations and Discussions} \label{sec:discussion}
In this section, we elaborate on insights encountered while conducting data collection, creating the prediction model, and analyzing the results.
Due to the COVID-19 pandemic, it was impossible to recruit a sufficient number and variety of participants. Whether the motion of people who have a stroke or have advanced age is the same as that of the participants should be considered in further investigations.
\hl{In \mbox{\cite{Coderre2010-fn, Scott2011-qh}}, the authors conducted a reaching-tasks experiment for mostly right-handed patients with hemiplegia.
They report that there is no difference in the motor function of the unaffected arm between left- and right-affected patients.
However, in comparisons between these patients and elderly individuals without paralysis, it is shown that patients with paralysis exhibited inferior motor function even in the unaffected arm.
}
Furthermore, \hl{In \mbox{\cite{Heller1987-af}}, it is shown that the arm motor functions of healthy elderly individuals differ depending on the dominant arm. 
From this, while it may seem difficult to directly apply the proposed model or the collected data from healthy individuals in this study to support a system intended for elderly or hemiplegic patients, the requirements for assistive systems identified through this study, along with the series of methods for model creation, can be considered useful.}
\hl{
Using the proposed model, it would be possible to realize a system where support robots autonomously operate triggered by the user's active movements, supporting the completion of user tasks.
In \mbox{\cite{Phillips2013-mp}}, the authors reported that there is a correlation between physical activity and self-efficacy, or life satisfaction in the elderly.
On the other hand, there have also been reported that elderly individuals have psychological barriers to engaging in physical activities in the first place (\mbox{\cite{Lee2008-ey}}).
Support systems in ADL can encourage active movements from familiar activities, as seen in this task, reduce psychological barriers to physical activity, and promote more extensive social activities. 
To achieve this, the support system aims to appropriately assist users' activities in daily life, enhancing their self-efficacy and motivation for active engagement. For this purpose, one of the future challenges is to enable support at multiple levels based on the user's physical condition, from simple arm support to higher-level assistance like directly retrieving objects and handing them to the user's extended hand.
}
\hl{
As mentioned in Section \mbox{\ref{sec:result}}, it is impossible to eliminate the possibility of misclassification when the classification model is used in the wild. While it is important to improve model performance, it is equally crucial to build and operate a system that is robust against misclassification.
The proposed model is expected to improve in accuracy with increased input frames (i.e., as the motion progresses). Additionally, the results from \mbox{\figurename{ \ref{fig:result}}} suggest that the proposed model achieves very high accuracy in the 3-class classification of rows (Left, Center, Right), with respective accuracies of 97.15\%, 96.60\%, and 93.47\%.
Hence, in the actual operation of support robots, for instance, at the beginning of a movement, the robot might perform lateral shifts, and as time progresses, it could execute more detailed movements based on predictive results. For interactions with people or objects, it would be practical to utilize proximity sensors equipped on the robot for precise positional adjustments.
}
However, it is difficult to say that the current model ensures sufficient time for actual support operations. As seen in Section \ref{sec:estimation_speed}, the time available for assistance is only about 0.96 s, which is clearly insufficient for a stationary robot to approach a user or target object and provide support.
Therefore, to achieve our goal, we must solve problems from many directions, such as establishing a fast support method, developing a soft robot that considers collision with humans or surrounding objects, and integrating these technologies, including this study. These are challenges for future study.

\section{Conclusions}\label{sec:conclusion}
We proposed a novel scheme for constructing a reaching-position prediction model for the reaching motion involving upper-arm lifting, which is part of activities of daily living (ADL), to develop a support robot.
Based on the results of the motion collection experiment and its analysis, we developed a target position prediction model using time-series data of face, motion, and depth features.
The proposed model, which demands that the support system autonomously operates triggered by the user's movements using data from simple sensors, achieved an accuracy of 93\% at 35\% of the motion completion time. This model, utilizing only 0.5 s of data, was able to make predictions in approximately 0.086 s of computation time.
\hl{
However, it is difficult to say that sufficient time has been secured for the operation of support robots,  and the issue of misclassification needs to be resolved to adapt the classification model in the wild.
In the future, along with improving prediction accuracy, we aim to develop robust support methods against misclassification and robots that support ADL in close contact with users, striving to realize the proposed system.
}

\section*{Declarations}
\subsection*{Availability of data and materials}
The datasets used and/or analysed during the current study are available from the corresponding author on reasonable request.
\subsection*{Competing interests}
The authors declare that they have no competing interests.
\subsection*{Funding}
This work is supported by JST [Moonshot R\&D],[Grant Number JPMJMS2034]
\subsection*{Ethical Approval and consent to participate}
Ethical approval was not required as per institutional guidelines.
All participants were informed about the purpose of the study, the anonymity and confidentiality of their results, and provided informed consent prior to participation.
\subsection*{Authors' contributions}
Y.T. worked on the research concept, participated in the design and development of the method, and drafted the thesis.
K.Y. participated in the research design. All authors reviewed the results and approved the final version of the manuscript. 
\subsection*{Acknowledgement}
Not applicable.

\bibliography{005_full_paper9}
\end{document}